\RequirePackage{fix-cm}
\documentclass[11pt,a4paper,onecolumn]{personalarxiv}

\usepackage{newfloat}
\usepackage{listings}
\usepackage{placeins}
\newtheorem{proposition}{Proposition}

\renewcommand{\headerfont}{\normalfont\color{black}\fontsize{8}{10}\selectfont}

\title{OrderGPT: Generation Order as a Learnable Modeling Choice}
\papershorttitle{OrderGPT: Learning the Generation Order}

\author[1]{Chenhe Yang}%
\author[1]{Yuhuan Lyu}
\author[1]{Tianyang Hu\thanks{Corresponding author:
  \href{mailto:hutianyang@cuhk.edu.cn}{\texttt{hutianyang@cuhk.edu.cn}}}}
\affil[1]{SAGA Lab, The Chinese University of Hong Kong, Shenzhen}
\paperpdfauthor{Chenhe Yang, Yuhuan Lyu, Tianyang Hu}

\begin{document}
\maketitle

\begin{abstract}

Generative modeling of discrete data has two popular paradigms: autoregressive models such as GPT, which factorize a sequence in a left-to-right (L2R) order, and masked diffusion models, which are essentially any-order autoregressive modeling. 
Rather than competing methods, these are the two extreme points of a single axis—the generation order. 
For structured discrete domains without a natural traversal, both extremes may be suboptimal, yet the intermediate choice—a single well-adapted order—remains largely unexplored.
We introduce \textit{OrderGPT}, which builds on Any-Order GPT to learn a \textit{task-level generation order} without canonical-order supervision.
Our key observation is that predicting variables from randomly sampled reveal prefixes produces recurring attention patterns in the backbone that provide a model-dependent signal of predictive dependencies.
We aggregate this signal into a reciprocal dependency graph and use a spectral teacher to derive continuous generation priorities.
A lightweight OrderMLP integrated into the GPT backbone learns to predict these priorities from attention, incorporating order learning into the architecture within a single training run. 
Training gradually shifts toward the learned order, which is then frozen for subsequent autoregressive training and generation.
The same mechanism adapts across modalities: adaptively it recovers near-L2R dependency structure in language, learns anatomically organized factorizations on human pose, and discovers a factorization on tabular data that places statistically and semantically related fields nearby, while improving downstream modeling metrics across all domains.
The code is available at \href{https://github.com/Chenhe-yang/OrderGPT/}{GitHub}.

\end{abstract}

\section{Introduction}
\label{sec:introduction}

Autoregressive generative models represent a joint distribution as a sequence of conditional prediction problems. This formulation underlies modern language models and has also been applied to visual tokens~\citep{esser2021taming}, human poses~\citep{lucas2022posegpt}, tabular data~\citep{yang2019deepcardinality}, graphs~\citep{you2018graphrnn}, and other structured domains~\citep{uria2016neural}. 
Although the chain rule permits the variables (or tokens) to be generated in any order, the choice of order matters for finite models: it determines the context available at each step, the difficulty of the conditional predictions, and the path along which errors propagate during generation~\citep{bengio2015scheduled}. 
Generation order is therefore an important modeling choice rather than merely an implementation detail.

Current methods largely occupy two endpoints of this design space. Conventional autoregressive models, such as GPT~\citep{radford2019language,brown2020language}, commit to a single prescribed factorization, typically the L2R order for language. When this order matches the structure of the data, it provides a strong inductive bias and allows the model to concentrate its capacity on a stable family of prediction problems.
Masked diffusion models (MDMs) take the opposite approach~\citep{lou2023discrete,shi2024simplified,sahoo2024simple,nie2026large,xie2026variational} by reconstructing masked variables from arbitrary observed subsets without specifying a canonical traversal. Their objective can be interpreted as any-order autoregressive training that averages over randomly sampled reveal permutations~\citep{ou2025your, xue2025any}. 
This flexibility is useful when no natural order exists, but it spreads a finite model's capacity across many factorizations of potentially unequal difficulty without identifying a stable order for training and generation.

\begin{figure}[t]
    \centering
    \vspace{-2mm}
    \includegraphics[width=1\textwidth]{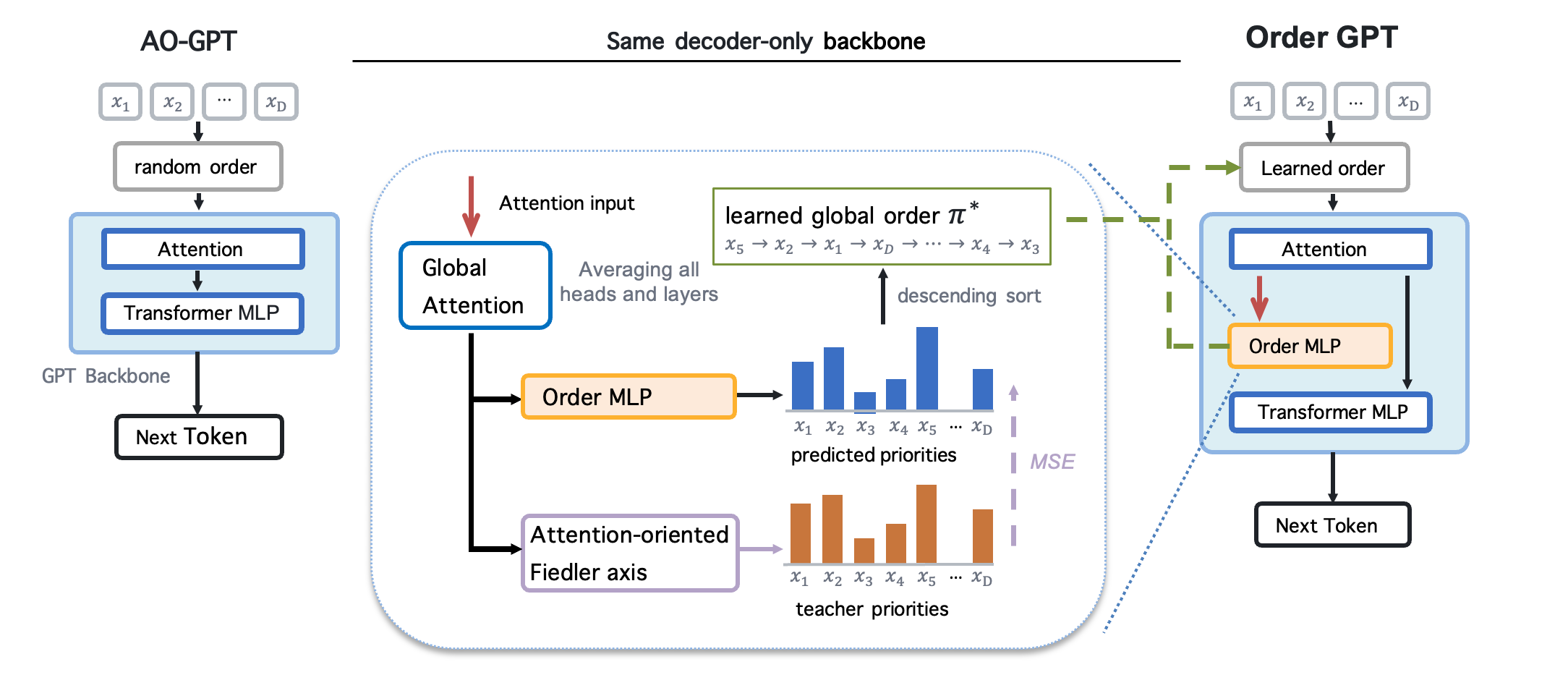}
    \vspace{-3mm}
    \caption{
       {OrderGPT augments the AO-GPT backbone with an Order MLP.
    The Spectral Teacher constructs a reciprocal affinity graph from aggregated AO-GPT attention, recovers a Fiedler axis, and orients it using directed attention.
    The resulting continuous priorities supervise the Order MLP, whose outputs are sorted to obtain the task-level generation order used by the backbone.}
       }
    \label{introduction}
\end{figure}

GPTs and MDMs can be viewed as two extremes along a common axis; both may be suboptimal for structured domains without a natural one-dimensional traversal. 
This motivates an intermediate alternative: \textit{learn a single task-level generation order and then use it in autoregressive factorization}. 
Learning such an order is nontrivial: the search space is factorial, canonical-order supervision is generally unavailable, and the most useful factorization may evolve with the model.

The decoder-only GPT architecture provides a natural backbone for this goal. 
Its computational architecture and its conventional L2R factorization are conceptually \textit{separable}: the same decoder can support different generation orders while retaining the GPT training and generation framework.
AO-GPT~\citep{xue2025any} makes this flexibility explicit by supporting arbitrary reveal orders within a decoder-only Transformer. 
It therefore provides both the architectural foundation for learning a generation order and an opportunity to observe predictive relationships across different reveal contexts.
Building on this backbone, we introduce \textbf{OrderGPT}, which incorporates a lightweight OrderMLP into AO-GPT to learn a task-level generation order without canonical-order supervision.
Figure~\ref{introduction} summarizes the OrderGPT pipeline.

Our key observation is that predicting the same variable from different randomly sampled prefixes produces recurring attention patterns that provide model-dependent evidence of predictive relationships. 
We aggregate these patterns to construct a reciprocal dependency graph.
Our \textit{Spectral Teacher} recovers a continuous axis from the graph Laplacian, placing strongly related variables nearby, then uses directed attention to orient it, favoring attended
context before its prediction targets.
The resulting continuous generation priorities supervise OrderMLP, which learns to predict them from aggregated attention across successive windows, consolidating recurring dependency evidence.
Sorting its predictions yields a task-level order that gradually replaces random orders during backbone training. The final order is frozen for subsequent autoregressive training and generation. 
OrderGPT thus integrates order learning into the GPT architecture within a single training run.

We evaluate OrderGPT across three domains with different structural properties.
In language, it learns an order close to L2R without L2R supervision and lowers perplexity relative to matched any-order
baselines.
On H3WB human-pose generation~\citep{zhu2023h3wb}, where no unique one-dimensional traversal exists, it learns anatomically coherent factorizations and improves density fitting and free-running generation. 
On the CENSUS tabular dataset~\citep{meek2001uscensus}, it places related fields nearby, improves density estimation, and reduces tail cardinality-estimation errors.

\section{Preliminaries}
\paragraph{Generation order as a modeling variable}
Let $\mathbf{x}=(x_1,\ldots,x_D)$ denote $D$ target variables,
$\mathbf{c}$ optional observed conditions, and
$\pi=(\pi_1,\ldots,\pi_D)$ a permutation of variable IDs.
Treating $\pi$ as an explicit modeling variable gives the
order-conditioned autoregressive factorization
$
  p_\theta(\mathbf{x}\mid\mathbf{c};\pi)
  =
  \prod_{t=1}^{D}
  p_\theta\!\left(
    x_{\pi_t}\mid x_{\pi_{<t}},\mathbf{c}
  \right).
$
$\sigma$-GPT~\citep{pannatier2024sigma} implements this formulation by providing positional encodings for both the input variable and the next variable to predict.
Causal attention follows the specified reveal order, while variable identities remain fixed across permutations.
This makes the generation order independently configurable within the same decoder architecture.

\paragraph{AO-GPT backbone}
\cite{xue2025any} provides an implementation of any-order GPT within a decoder-only Transformer architecture. 
Specifically, it injects the target-position embedding into every Transformer block through adaptive layer normalization (adaLN).
The embedding determines separate scale, shift, and residual-gating parameters for the attention and feed-forward branches, allowing the prediction target to modulate computation throughout the network.
The backbone retains causal attention along the reveal order.
For a specified permutation, it supports likelihood evaluation in a single forward pass and key--value caching during sequential generation. 
Given a distribution $q_{\mathrm{AO}}$ over reveal permutations,
the corresponding order-averaged objective is
\begin{equation}
  \mathcal{L}_{\mathrm{AO}}(\theta)
  =
  \mathbb{E}_{(\mathbf{x},\mathbf{c})}
  \mathbb{E}_{\pi\sim q_{\mathrm{AO}}}
  \left[
  -\sum_{t=1}^{D}
  \log p_\theta\!\left(
    x_{\pi_t}\mid x_{\pi_{<t}},\mathbf{c}
  \right)
  \right].
  \label{eq:any-order-objective}
\end{equation}
In the standard any-order setting, $q_{\mathrm{AO}}$ is uniform over all $D!$ permutations.
For discrete data, the objective in Eq.~\eqref{eq:any-order-objective} is equivalent to the standard masked diffusion training objective at the formulation level~\citep{xue2025any,ou2025your}.

\paragraph{Learning generation orders}
In encoder-only MDM literature, it is known that training contains difficult infilling subproblems and that a suitable decoding order exists~\citep{pmlr-v267-kim25ah,peng2025path,garg2025masked,hong2026unifying}.
A closely related approach is LO-ARM~\citep{pmlr-v267-wang25cn}, which treats the ordering as latent variables and learns a state-dependent order policy that sequentially selects the next variable. 
The policy is trained jointly with the autoregressive model through a variational lower bound on the log-likelihood.
In comparison, OrderGPT learns a single task-level permutation $\pi^*$ that captures dependency structure shared across samples.  It recovers this permutation from attention in an AO-GPT backbone, without introducing a sequential order policy or optimizing a distribution over orders. 
Once learned, the global order is frozen and reused for ordinary autoregressive training and generation, providing an explicit structural inductive bias without per-instance order search.

\begin{figure}[!t]
  \centering
  \includegraphics[
    width=1\textwidth
  ]{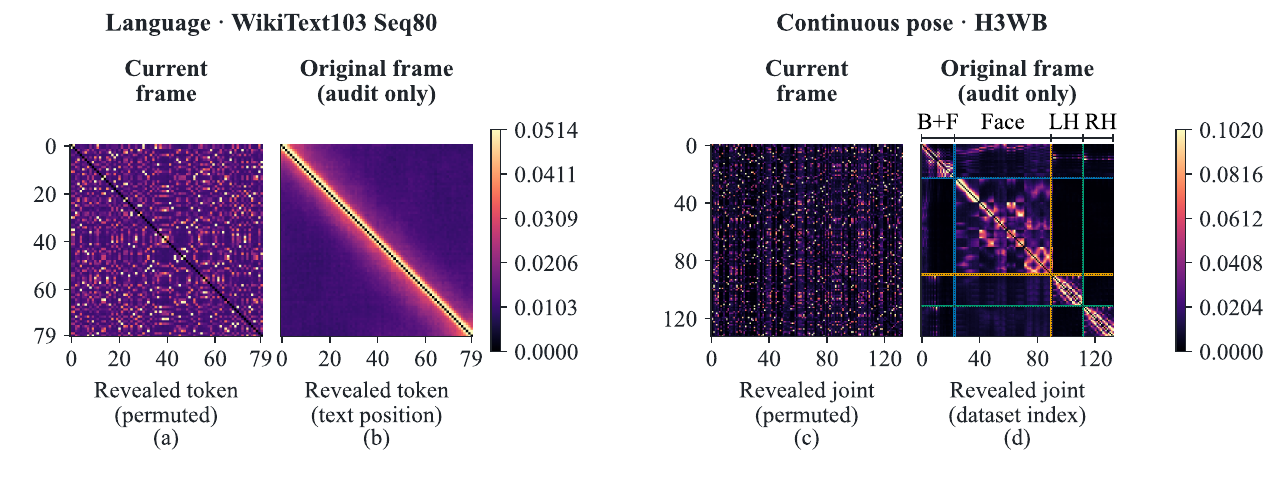}
  \vspace{-7mm}
 \caption{Attention patterns under the current and original indexing.
 \textbf{(a, b)} WikiText103 Seq80 attention.
 \textbf{(c, d)} H3WB attention.
 The original-frame panels reorder the same values only for visualization, highlighting local tokens and anatomical structure. Each current/original pair shares a color scale.}
  \label{fig:attention-current-original}
\end{figure}

\section{Attention Reveals Order Preference}
\label{sec:method}
Generation order controls the context available to each prediction.
We first show how this choice can affect the conditional distributions that a model with limited output capacity can represent.
We then use attention from random-order training to estimate relationships between variables.
Finally, we convert these relationships into a generation order.

\subsection{ Order Matters under Limited Capacity}
\label{sec:factorization-orders}

The chain rule allows any generation order, but different orders require different conditional predictions.
We formalize the resulting capacity tradeoff in a toy setting, showing that a suitable order can achieve a strictly lower population loss than uniform random-order training at the same model capacity.

We consider categorical variables forming a Markov chain along a dependency path $\rho$. 
This path describes the data's dependencies and need not match the original variable indexing. 
To make the capacity constraint explicit, all predictions use a shared categorical output layer with $C$ features.
Each prediction can still condition on the entire revealed prefix.
We measure performance by expected negative log-likelihood per variable.
Following the chain makes a neighboring variable available for each prediction after the first, so the model only needs to represent the local one-step conditionals. 
An order that skips an intermediate variable can instead require a longer-range conditional obtained by averaging over unrevealed states.
Our construction makes the local conditionals representable within the shared output capacity, while a skipped-intermediate conditional has unavoidable approximation error.

\begin{proposition}[Informal]
\label{prop1}
Let $\mathcal L_{\mathrm{chain}}^\star(C)$ denote the best achievable loss under the chain order, and let $\mathcal L_{\mathrm{random}}^\star(C)$ denote the best average loss achievable by one predictor across uniformly sampled orders, within the same predictor family.
There exist full-support reversible categorical Markov chains
and capacity limits $C$ for which the true joint distribution
is exactly representable within this predictor family under
the chain order and its reversal, whereas
\begin{equation}
  \mathcal L_{\mathrm{random}}^\star(C)
  \ge
  \mathcal L_{\mathrm{chain}}^\star(C)+\Delta_C,
  \qquad \Delta_C>0.
  \label{eq:order-capacity-gap}
\end{equation}
\end{proposition}
Uniform random ordering encounters the difficult prediction problems with positive probability, producing a non-vanishing
gap in the best achievable population loss. 
At the stated capacity, this gap persists even with unlimited data and optimal fitting. Appendix~\ref{app:order-capacity-proof} gives the precise model definition, capacity conditions, construction, and proof.

The chain order and its reversal are effective because they make an informative neighboring variable available for each prediction after the first. 
This example motivates selecting orders that supply informative context for later predictions.
In real data, the relevant relationships are unknown; we next investigate attention as a model-dependent signal for identifying them.

\subsection{From Attention to a Reciprocal Dependency Graph}
\label{sec:affinity-estimation}

To identify predictive relationships that can guide order learning, we examine attention in an AO-GPT backbone trained with random reveal orders under Eq.~\eqref{eq:any-order-objective}.
These orders expose each target to different prefixes, allowing us to examine which contextual variables repeatedly receive attention when available.

At the start of each run, we randomly relabel the variables and keep these IDs fixed across examples.
Original IDs are used only for post-hoc evaluation.
Figure~\ref{fig:attention-current-original} compares the same attention values under relabeled and original variable IDs.
Restoring the original indexing reveals a diagonal band in language, indicating stronger attention between nearby text positions.
In H3WB, distinct attention blocks align with body/feet, face, and the left and right hands.
These observations motivate using attention as a model-dependent signal of relationships between variables.

We aggregate attention across training examples and random generation orders into a directed matrix $A$.
Each entry $A_{ij}$ averages attention to variable $j$ when predicting variable $i$, using only predictions where $j$ is already observed.
To capture relationships supported in both directions, we construct a symmetric reciprocal dependency graph:
\begin{equation}
  W_{ij}=\sqrt{A_{ij}A_{ji}}\quad(i\ne j),
  \qquad W_{ii}=0.
  \label{eq:reciprocal-affinity}
\end{equation}
The geometric mean favors mutually strong attention over one-sided links. 
The reciprocal graph $W$ captures which variables are strongly related, while the directed matrix $A$ preserves information about which variables are attended to when predicting others. We use these complementary signals to construct and orient a generation order.

\subsection{Laplacian Recovery and Attention-Based Orientation}
\label{sec:reciprocal-spectral-recovery}

We now turn the reciprocal dependency graph $W$ into a generation order. 
Motivated by the dependency-following orders in Section~\ref{sec:factorization-orders}, we seek an arrangement that places strongly connected variables nearby. 
We assign each variable a coordinate on a line and use a graph-Laplacian objective to penalize separation between connected variables. Sorting these coordinates yields a candidate order; we then use directed attention to choose its orientation.

Let $D_W$ be the diagonal matrix with $(D_W)_{ii}=\sum_j W_{ij}$.
The graph Laplacian is $L=D_W-W$.
Let $\mathbf z$ collect the variable coordinates, with $z_i$ assigned to variable $i$.
We measure the separation between connected variables by
\begin{equation}
  \mathcal E_W(\mathbf z)
  =\mathbf z^{\mathsf T}L\mathbf z
  =\frac12\sum_{i,j}W_{ij}(z_i-z_j)^2.
  \label{eq:laplacian-locality}
\end{equation}
We minimize this energy while fixing the coordinates to have zero mean and unit length~\citep{fogel2015convex}:
\begin{equation}
  \begin{aligned}
    \mathbf v_2
      &\in\arg\min_{\sum_i z_i=0,\,
                        \|\mathbf z\|_2=1}
                  \mathbf z^{\mathsf T}L\mathbf z,\\
    \pi^{\mathrm{raw}}
      &=\operatorname{argsort}(\mathbf v_2).
  \end{aligned}
  \label{eq:fiedler-axis}
\end{equation}
Sorting the entries of $\mathbf v_2$ in ascending order gives the candidate generation order $\pi^{\mathrm{raw}}$.
For a connected graph, $\mathbf v_2$ is a Fiedler vector, associated with the second-smallest eigenvalue of $L$~\citep{atkins1998spectral}.
When the second-smallest eigenvalue is not repeated and all coordinates are distinct, the order is unique up to reversal.

We use the directed attention matrix $A$ to choose between these two directions.
For an order $\pi$, let $S(\pi)$ be the total attention from later variables to earlier ones:
\begin{equation}
  \begin{aligned}
    S(\pi)&=\sum_{1\le s<t\le D}A_{\pi_t,\pi_s},\\
    \pi^{\mathrm{spec}}&\in\arg\max_{\pi\in
      \{\pi^{\mathrm{raw}},
        \operatorname{reverse}(\pi^{\mathrm{raw}})\}}S(\pi).
  \end{aligned}
  \label{eq:attention-orientation}
\end{equation}
After reordering $A$ by $\pi^{\mathrm{raw}}$, the forward score is the sum below the diagonal, and the reverse score is the sum above it.
We select the higher-scoring direction as $\pi^{\mathrm{spec}}$.
Equal scores indicate no directional preference.
This choice uses only the training attention matrix.
Appendix~\ref{app:attention-orientation} provides the derivation.
We orient the Fiedler coordinates so that variables appearing earlier in $\pi^{\mathrm{spec}}$ have larger values.
We call the complete procedure the \emph{Spectral Teacher}.

\paragraph{Order recovery}
For the Markov-chain example in Section~\ref{sec:factorization-orders}, suppose the attention graph is sufficiently close to a graph connecting neighboring variables along the chain.
Spectral sorting then recovers the chain order up to reversal.
Both directions retain the advantage over random orders at the same output capacity.
A trained model can achieve this loss advantage if it learns the corresponding conditional predictions accurately enough.
Appendix~\ref{app:spectral-recovery-proof} gives the precise conditions and proof.
On WikiText103 Seq80, the Spectral Teacher recovers an order close to L2R, with Kendall's $\tau=0.989$.
L2R is used only as a reference for evaluating the recovered order.
Section~\ref{sec:integrated-training} compares the Spectral Teacher's predictive performance and computational cost with those of the Order MLP.

\section{OrderGPT}
\label{sec:ordergpt}
Section~\ref{sec:method} shows that recurring attention patterns in AO-GPT provide a model-dependent signal of predictive dependencies, which the Spectral Teacher converts into continuous generation priorities. 
We now incorporate this supervision into a trainable component of the GPT architecture, so that the model can learn its factorization as the backbone develops and use it to guide subsequent training within the same run.

\begin{figure}[!t]
  \centering
  \vspace{-4mm}
  \includegraphics[
    width=1\textwidth
  ]{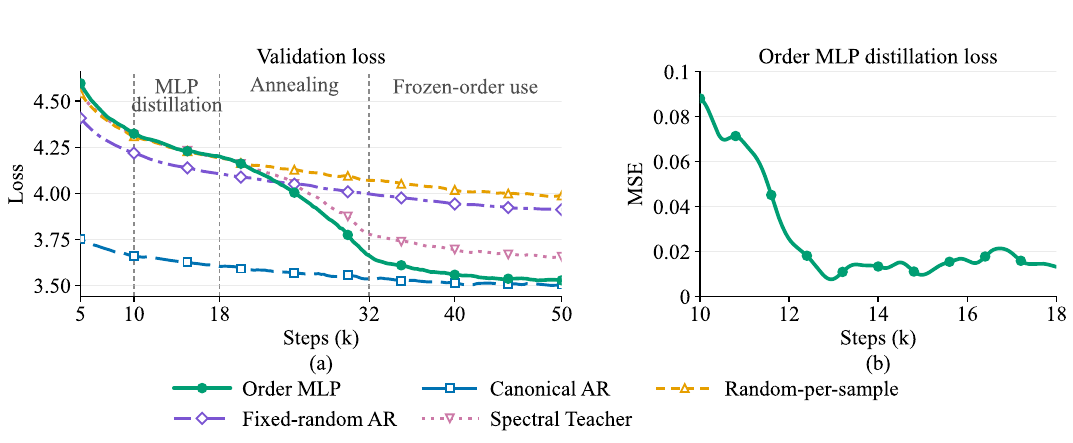}
  \vspace{-4mm}
  \caption{Order MLP effectiveness on WikiText103 Seq80.
  \textbf{(a)} Validation loss under Order MLP, Spectral Teacher, Random-per-sample, Fixed-random AR, and Canonical AR.
  Vertical guides mark MLP distillation (10k--18k), annealing of the learned-order (18k--32k), and full use of the frozen learned order after 32k.
  The final learned axis has a positive post-hoc original-frame Kendall's $\tau=0.999$.
  \textbf{(b)} Distillation MSE between Order-MLP priorities and oriented Fiedler targets. }
  \label{fig:teacher-mlp-distillation}
\end{figure}

\subsection{Order MLP Distillation}
\label{sec:order-mlp}

We augment the AO-GPT backbone with a lightweight OrderMLP that maps the directed attention matrix $A$ to one scalar priority per variable, with larger values indicating earlier generation.
Training a shared mapping across successive attention windows encourages the module to capture recurring dependency structure and reduces its reliance on individual spectral estimates.

\paragraph{Supervision and training objective} 
The Spectral Teacher rescales the Fiedler coordinates, oriented by Eq.~\eqref{eq:attention-orientation}, to $[0,1]$, assigning larger priorities to earlier variables.
We train OrderMLP by minimizing the mean squared error between predicted and teacher priorities.
The attention input is detached, so this loss updates only OrderMLP. 
Sorting its predictions in descending order yields the candidate generation order $\hat\pi_{\mathrm{MLP}}$.

\subsection{Integrated Training Recipe}
\label{sec:integrated-training}
As shown in Figure~\ref{introduction}, AO-GPT initially trains with random reveal orders.
During this phase, the Spectral Teacher derives generation priorities from AO-GPT's attention.
The Order MLP learns to predict these priorities and converts them into a candidate generation order.
AO-GPT continues to use random reveal orders throughout this initial distillation phase.

We then gradually increase the probability of training AO-GPT with the learned order, while sampling random orders otherwise.
During the configured update window, the Order MLP can refresh the candidate order as AO-GPT's attention evolves.
At the end of this window, we freeze the final accepted order as $\pi^*$.
AO-GPT reuses this order across examples for subsequent autoregressive training and generation.

\paragraph{Effectiveness}
Figure~\ref{fig:teacher-mlp-distillation}(a) shows that Order MLP attains a lower final validation loss than Spectral Teacher, and closely approaches Canonical AR.
The learned axes have positive post-hoc Kendall's $\tau$ compared with natural L2R: $\tau=0.989$ for Spectral Teacher and $\tau=0.999$ for Order MLP.
Figure~\ref{fig:teacher-mlp-distillation}(b) provides complementary evidence: after an initial transient, the distillation MSE decreases and remains low across successive Spectral Teacher windows.
Because the same MLP is optimized across these windows, its shared parameters consolidate recurring structure rather than copying each small, window-specific rank fluctuation into the executable order.
This helps explain the lower validation loss and stronger final order agreement in panel a. 

\paragraph{Efficiency}
We isolate order production with an optimized fixed-checkpoint replay at the matched Seq80 refresh cadence.
The Spectral Teacher's spectral path requires $O(D^2)$ storage for the graph and $O(D^3)$ time for the dense eigendecomposition; the frozen MLP consumes the already aggregated attention matrix and replaces the repeated spectral solve with a forward pass and sorting.
In the replay, the Spectral Teacher requires $242.727$ ms per optimizer iteration, whereas the frozen Order MLP requires $190.992$ ms.
After normalization by their paired Random-per-sample baselines, the per-iteration replay time is $21.336\%$ lower for the Order MLP than for the Spectral Teacher.

\begin{table}[t]
  \centering
  \small
  \caption{Zero-shot perplexity ($\downarrow$) on language benchmarks for approximately 125M-parameter models.
  Published baselines are reproduced from Table~5 of AO-GPT~\citep{xue2025any}.
  $^\dagger$ denotes a model reproduced by the AO-GPT authors, and $^\ddagger$ denotes AO-GPT trained with an additional 10\% of L2R-ordered data.
  $^\ast$ denotes our AO-GPT baseline trained for 400k updates without language-model EMA or a prescribed L2R training mixture.
  Published rows follow their original settings and provide numerical context rather than a compute-matched ranking.
  Fixed-random AR (throughout) uses one presampled random permutation throughout training and inference.
  Fixed-random AR + curriculum follows OrderGPT's random-order warm-up and order-use annealing schedule, with a presampled random permutation as its terminal order for subsequent training and inference.
  }
  \label{tab:openwebtext-language-draft}
  \setlength{\tabcolsep}{6pt}
  \begin{tabular}{@{}lrrrrr@{}}
    \toprule
    \footnotesize
    Model and inference order & LAMBADA & WikiText-2 & PTB & WikiText-103 & 1BW \\
    \midrule
    \multicolumn{6}{@{}l}{\emph{Published L2R evaluation}} \\
    SEDD & 49.410 & 41.190 & 118.740 & 41.700 & 72.600 \\
    RADD & 49.090 & 38.260 & 107.780 & 38.410 & 63.330 \\
    GPT-2 & 45.040 & 42.430 & 138.430 & 41.600 & 75.200 \\
    $\sigma$-GPT$^\dagger$ & 68.610 & 57.660 & 146.870 & 55.540 & 90.980 \\
    AO-GPT$^\ddagger$ & 52.460 & 42.100 & 135.960 & 40.970 & 71.730 \\
    \midrule
    \multicolumn{6}{@{}l}{\emph{Published order-agnostic evaluation}} \\
    SEDD & 50.920 & 41.840 & 114.240 & 40.620 & 79.290 \\
    RADD & 50.270 & 38.260 & 110.380 & 35.900 & 74.280 \\
    $\sigma$-GPT$^\dagger$ & 65.830 & 53.080 & 138.610 & 50.750 & 87.710 \\
    AO-GPT$^\ddagger$ & 59.930 & 46.330 & 141.920 & 45.440 & 84.360 \\
    AO-GPT$^\ddagger$ + Ensemble (8) &
    55.620 & 42.770 & 126.080 & 42.020 & 77.620 \\
    AO-GPT$^\ddagger$ + Ensemble (64) &
    54.920 & 42.240 & 123.490 & 41.510 & 76.730 \\
    \midrule
    \multicolumn{6}{@{}l}{\emph{Our 400k-update evaluation (no language-model EMA)}} \\
AO-GPT$^\ast$ + Random inference &
69.345 & 57.606 & 161.210 & 55.635 & 90.168 \\
AO-GPT$^\ast$ + Canonical L2R inference &
70.262 & 60.273 & 163.053 & 58.535 & 91.069 \\
Fixed-random AR (throughout) &
71.140 & 57.830 & 160.790 & 56.310 & 90.390 \\

Fixed-random AR + curriculum &
67.690 & 53.600 & 149.530 & 51.800 & 86.870 \\
\textbf{OrderGPT (Order MLP)} &
\textbf{50.962} & \textbf{45.125} & \textbf{132.753} &
\textbf{39.217} & \textbf{70.046} \\
    \bottomrule
  \end{tabular}
\end{table}

\section{Empirical Validation}
\label{sec:experiments-setup-draft}

We evaluate OrderGPT on language modeling, continuous human-pose generation, and categorical tabular modeling.
These domains cover discrete sequences, structured continuous variables, and variables without a prescribed topology.
We test whether learning task-level order improves performance over random-order training.
Our main comparison is between OrderGPT and AO-GPT.
Within each domain, the two methods use the same backbone capacity, data processing, and training budget.

Canonical AR uses L2R in language and the original dataset indexing in H3WB and CENSUS.
Fixed-random AR tests whether an arbitrary fixed order provides similar gains.
In language, we evaluate the same AO-GPT$^\ast$ model with random and L2R inference orders to test whether changing inference order alone reproduces the gains.
Fixed-random AR + curriculum follows OrderGPT's warm-up and annealing schedule but ends with a presampled random order, testing whether the schedule alone reproduces the gains.

\subsection{Language Modeling}
\label{sec:experiments-language-draft}

We train an approximately 125M-parameter decoder-only model on OpenWebText with a context length of 1024, and align the five-dataset evaluation suite with Table~5 of AO-GPT~\citep{xue2025any}.

\paragraph{Learned order}
The final learned order closely matches L2R, with Kendall's $\tau=0.997$ after mapping variables back to their original positions.
L2R is used only for this post-hoc comparison and does not guide order learning.

\begin{table}[t]
  \centering
  \small
\caption{Performance on H3WB masked-I2D.
Internal results are reported as mean $\pm$ sample standard deviation across three independently trained model seeds ($n=3$ per method).
Each run is evaluated on the same 10,000-pose test split.
NLL is averaged per joint, and MPJPE values are in millimeters.
Published results are taken from Table~4 of \cite{zhu2023h3wb} and provide external numerical context.
Lower is better for every metric.}
  \label{tab:h3wb-masked-i2d-draft}
  \setlength{\tabcolsep}{4.5pt}
  \footnotesize
  \begin{tabular}{@{}llccccc@{}}
    \toprule
    Method & Training data & NLL/joint $\downarrow$ & Whole $\downarrow$ &
    Body $\downarrow$ & Face-local $\downarrow$ & Hand-local $\downarrow$ \\
    \midrule
    Large SimpleBaseline & H3WB & -- & 131.40 & 131.60 & 19.80 & 44.80 \\
    Jointformer & H3WB & -- & 109.20 & 103.00 & 19.80 & 53.50 \\
    Jointformer & H3WB+T3WB & -- & 84.20 & 80.10 & 16.30 & 44.50 \\
    \midrule
Canonical AR & H3WB & $-11.05 \pm 0.02$ & $80.83 \pm 1.80$ &
$74.86 \pm 0.59$ & $10.87 \pm 0.66$ & $37.41 \pm 3.87$ \\
Fixed-random AR & H3WB & $-10.68 \pm 0.09$ & $99.26 \pm 11.87$ &
$82.69 \pm 8.53$ & $13.14 \pm 2.68$ & $36.36 \pm 1.71$ \\
AO-GPT & H3WB & $-10.59 \pm 0.07$ & $115.23 \pm 15.98$ &
$89.97 \pm 8.40$ & $14.67 \pm 2.95$ & $38.18 \pm 0.42$ \\
\textbf{OrderGPT} & H3WB &
$\boldsymbol{-11.30 \pm 0.04}$ &
$\boldsymbol{79.44 \pm 1.76}$ &
$\boldsymbol{72.40 \pm 0.73}$ &
$\boldsymbol{10.16 \pm 0.74}$ &
$\boldsymbol{30.03 \pm 0.52}$ \\
    \bottomrule
  \end{tabular}
\end{table}

\begin{table}[t]
  \centering
  \small
\caption{Performance in density modeling and cardinality estimation on CENSUS.
Internal results are reported as mean $\pm$ sample standard deviation across three independently trained model seeds ($n=3$ per method).
Test NLL is averaged per variable on the held-out split and displayed in units of $10^{-3}$.
Published rows reproduce the reported 10,000-particle results~\citep{yang2019deepcardinality, liang2020variableskipping}.
Internal Q-error uses the same 1,000-query workload with 10,000 particles per query.
Lower is better, and bold denotes the lowest internal mean.}
  \label{tab:census-density-cardinality-draft}
  \setlength{\tabcolsep}{6pt}
  \footnotesize
  \begin{tabular}{@{}lcccc@{}}
    \toprule
    Method & Test NLL/variable ($\times 10^{-3}$) $\downarrow$ & Q-error P50 $\downarrow$ &
    Q-error P99 $\downarrow$ & Maximum Q-error $\downarrow$ \\
    \midrule
    \multicolumn{5}{@{}l}{\emph{Published references}} \\
    Progressive Sampling & -- & $1.030 \pm 0.010$ &
    $1.490 \pm 0.080$ & $8.860 \pm 14.000$ \\
    Variable Skipping & -- & $1.090 \pm 0.010$ &
    $2.480 \pm 0.290$ & $14.700 \pm 7.000$ \\
    Multi-Order + Skipping & -- & $1.200 \pm 0.240$ &
    $3.200 \pm 0.290$ & $12.600 \pm 5.000$ \\
    \midrule
    \multicolumn{5}{@{}l}{\emph{Internal evaluations}} \\
    Canonical AR & $325.818 \pm 0.020$ & $1.029 \pm 0.003$ &
    $1.459 \pm 0.007$ & $2.862 \pm 1.137$ \\
    Fixed-random AR & $325.983 \pm 0.019$ & $1.032 \pm 0.004$ &
    $1.514 \pm 0.100$ & $4.844 \pm 0.751$ \\
    AO-GPT & $332.665 \pm 0.148$ & $1.046 \pm 0.012$ &
    $1.671 \pm 0.006$ & $10.500 \pm 4.478$ \\
    \textbf{OrderGPT} & $\boldsymbol{325.488 \pm 0.073}$ &
    $\boldsymbol{1.027 \pm 0.001}$ &
    $\boldsymbol{1.442 \pm 0.013}$ &
    $\boldsymbol{2.333 \pm 1.014}$ \\
    \bottomrule
  \end{tabular}
  \vspace{2pt}

  \parbox{\textwidth}{\footnotesize\emph{Note.}
  For the positive cardinalities in this workload, with estimated cardinality $\hat c$ and true cardinality $c$, $\mathrm{Q\text{-}error}=\max(\hat c/c,c/\hat c)$; its ideal value is one.}
\end{table}

\paragraph{Improved performance}
For zero-shot benchmarking, OrderGPT and our AO-GPT$^\ast$ baseline are trained for 400k updates.\footnote{
Both runs match the 400k-update budget of AO-GPT$^\ddagger$ but do not use language-model EMA or its 10\% L2R training mixture.
}
Following AO-GPT, we report zero-shot perplexity on LAMBADA, WikiText-2, Penn Treebank (PTB), WikiText-103, and the One Billion Word Benchmark (1BW).
We evaluate the same trained AO-GPT$^\ast$ model under either random inference orders or canonical L2R inference, whereas OrderGPT uses its frozen task-level generation order.
As shown in Table~\ref{tab:openwebtext-language-draft}, OrderGPT achieves lower perplexity than all internal baselines on all five benchmarks.
For AO-GPT$^\ast$, L2R inference yields higher perplexity than random inference on every benchmark.
Changing the inference order alone therefore does not reproduce OrderGPT's gains.
Fixed-random AR + curriculum improves over both AO-GPT$^\ast$ with random inference and Fixed-random AR (throughout), but remains worse than OrderGPT on all five benchmarks.
Thus, the gains observed with OrderGPT are not fully reproduced by a curriculum that terminates in an arbitrary fixed order.

\subsection{Human-Pose Generation}
\label{sec:experiments-h3wb-draft}

The H3WB masked-I2D task~\citep{zhu2023h3wb} requires the model to generate a complete 3D whole-body pose from incomplete 2D joint observations.
Because human joints have local anatomical dependencies but no uniquely correct one-dimensional traversal, the task tests whether OrderGPT can learn a useful continuous-variable factorization without using the original joint numbering.

\paragraph{Learned order}
After mapping the final learned orders back to the original joint coordinates, we observe a consistent coarse anatomical-region traversal of face $\rightarrow$ body/feet $\rightarrow$ hands.
This recurring pattern differs from the canonical H3WB indexing, which follows body/feet $\rightarrow$ face $\rightarrow$ hands.
The shared region-level pattern suggests an ordering preference specific to the model and task.

\paragraph{Improved performance}
We next test whether the complete learned-order pipeline improves both density fitting and closed-loop pose generation.
Teacher-forced negative log-likelihood (NLL) measures density fitting, whereas mean per-joint position error (MPJPE) measures 3D coordinate accuracy after a complete autoregressive rollout.\footnote{Whole and Body MPJPE use pelvis alignment, Face-local MPJPE uses nose alignment, and Hand-local MPJPE uses wrist alignment.}
Averaged over the reported runs, OrderGPT achieves the lowest reported mean NLL and mean MPJPE in all four MPJPE columns among the internal methods in Table~\ref{tab:h3wb-masked-i2d-draft}.
These results reflect the complete learned-order pipeline, while Canonical AR already contains an anatomical-layout prior through the H3WB joint numbering.

\subsection{Tabular Estimation}
\label{sec:experiments-census-draft}

CENSUS~\citep{meek2001uscensus} is a categorical tabular dataset where table columns have no natural generation order, making this a test of learning an order from their statistical dependencies.
Alongside density modeling, we evaluate cardinality estimation, which predicts how many table rows satisfy a query.

\paragraph{Learned order}
Across independent runs, variables with strong mutual attention tend to appear close together in the learned order.
We quantify this pattern using a locality score that measures average separation in generation positions.
Lower scores indicate closer placement.
For a representative run, the score is 17.910 for the learned order, 23.170 for the canonical column order, and 23.000 in expectation under random ordering.
Original column identities and names are used only for this post-hoc interpretation.

\paragraph{Improved performance}
We measure density modeling with mean test NLL and cardinality estimation with Q-error.
Q-error measures the factor by which a query's row count is overestimated or underestimated.
We evaluate 1,000 queries using 10,000 progressive-sampling particles per query.
The median (P50) summarizes typical errors, while the 99th percentile (P99) and maximum capture larger errors.
OrderGPT has the lowest reported mean on every metric among the internal methods in Table~\ref{tab:census-density-cardinality-draft}.
Relative to AO-GPT, the largest reduction among the Q-error statistics is in the mean maximum.
Compared with Canonical AR, the differences are smaller, particularly for median Q-error.
Thus, the complete OrderGPT pipeline improves both density modeling and cardinality estimation over AO-GPT, with particularly large gains in reducing the largest query errors.

\section{Discussion}
\label{sec:discussion}

Our results support treating generation order as a learnable model choice within decoder-only autoregressive models.
OrderGPT integrates attention-based order learning into the GPT backbone, learning a task-level factorization automatically without canonical-order supervision.
Across language, human pose, and tabular data, the learned orders capture modality-specific structure and improve downstream performance, suggesting that a learned, stable factorization provides a useful middle ground between prescribed orders and uniform random-order training.

The current method shares one order across all examples, making it most suitable for tasks with recurring dependency structure.
A global order may become restrictive when dependencies vary substantially across samples, and the attention signal itself remains conditioned on the model and reveal policy.
Future work could combine reusable global order structure with context-dependent selection based only on variables available during generation.

\bibliographystyle{plainnat}
\bibliography{reference}

\newpage

\clearpage
\appendix
\setcounter{figure}{0}
\setcounter{table}{0}
\renewcommand{\thefigure}{S\arabic{figure}}
\renewcommand{\thetable}{S\arabic{table}}
\numberwithin{equation}{section}

\begin{center}
{\Large\bfseries Supplementary Material}\par
\medskip
{\large OrderGPT: Generation Order as a Learnable Modeling Choice}
\end{center}

\section{Theoretical Analysis and Complete Proofs}
\label{app:theory}

We prove an order gap for a specified categorical output class, then give sufficient conditions connecting its beneficial axis to the Spectral Teacher and Order MLP.
All logarithms are natural, and risks are negative log-likelihoods per variable.

\subsection{Setup and finite-capacity risk}
\label{app:theory-setup}

Let $P$ be a strictly positive distribution on $\{1,\ldots,V\}^D$, and let $\mathfrak S_D$ be the permutations of the $D$ variable IDs.
The history $h_{\pi,t}$ contains target ID $\pi_t$ and the complete ordered prefix with its values and IDs.
Consider the shared output class
\begin{equation}
\mathcal F_C=
\left\{q(a\mid h)=\operatorname{softmax}\bigl(U\phi(h)+b\bigr)_a:
U\in\mathbb R^{V\times C},\ b\in\mathbb R^V\right\},
\label{app:theory-model-class}
\end{equation}
where $U,b$ are shared across all histories, and $\phi$ may be any finite-valued function into $\mathbb R^C$.
Thus $C$ limits the shared logit-feature dimension, while the complete prefix remains available.
This is an output-interface capacity model related to the softmax bottleneck viewpoint.\footnote{Zhilin Yang, Zihang Dai, Ruslan Salakhutdinov, and William W. Cohen, \emph{Breaking the Softmax Bottleneck: A High-Rank RNN Language Model}, ICLR 2018, \url{https://arxiv.org/abs/1711.03953}.}
The order-separation construction is proved below.

Define
\begin{equation}
\begin{aligned}
Q_{q,\pi}(x)&=\prod_{t=1}^D q(x_{\pi_t}\mid h_{\pi,t}),\\
R(q,\pi)&=-\frac1D\mathbb E_P\log Q_{q,\pi}(X)
=\frac{H(P)}D+\frac1D\operatorname{KL}(P\Vert Q_{q,\pi}).
\end{aligned}
\label{app:theory-risk-kl}
\end{equation}
The last equality follows from the chain rule and cross-entropy identity; unrestricted exact conditionals give $H(P)/D$ for every order.
The fixed-order and uniform any-order optima are
\begin{equation}
\begin{aligned}
R_C^\star(P,\pi)&=\inf_{q\in\mathcal F_C}R(q,\pi),\\
R_{\mathrm{AO},C}^\star(P)&=
\inf_{q\in\mathcal F_C}
\mathbb E_{\Pi\sim\operatorname{Unif}(\mathfrak S_D)}R(q,\Pi).
\end{aligned}
\label{app:theory-ao-optimum}
\end{equation}
The latter infimum remains outside the expectation because one predictor is shared across orders.
It corresponds to Eq.~\eqref{eq:any-order-objective} divided by $D$, specialized to uniform orders and no additional conditioning variables.
For the chain construction below, the main-text quantities satisfy $\mathcal L_{\mathrm{chain}}^\star(C)=R_C^\star(P,\rho)$ and $\mathcal L_{\mathrm{random}}^\star(C)=R_{\mathrm{AO},C}^\star(P)$.

\subsection{A strict order gap under a shared softmax bottleneck}
\label{app:order-capacity-proof}

\paragraph{Theorem A.1 (finite-capacity order separation).}
For every $D\ge3$, $V\ge4$, and $2\le C\le V-2$, there exist a strictly positive $P$ and an axis $\rho\in\mathfrak S_D$ such that $Y_t=X_{\rho_t}$ is a stationary reversible first-order Markov chain and
\begin{equation}
\begin{aligned}
R_C^\star(P,\rho)
&=R_C^\star(P,\operatorname{rev}(\rho))=\frac{H(P)}D,\\
R_{\mathrm{AO},C}^\star(P)&\ge\frac{H(P)}D+\Gamma_C,
\qquad \Gamma_C>0.
\end{aligned}
\label{app:theory-capacity-theorem}
\end{equation}
The two chain directions require at most two shared features for exact representation, whereas an order beginning with $(\rho_1,\rho_3)$ requires at least $V-1$.

\paragraph{Proof: construction and the easy directions.}
For distinct real $s_1,\ldots,s_V$ and finite $\beta_1,\ldots,\beta_V$, set
\begin{equation}
\begin{aligned}
Z_a&=\sum_{b=1}^V\exp(\beta_b+s_as_b),
&T_{ab}&=\frac{\exp(\beta_b+s_as_b)}{Z_a},\\
\nu_a&=\frac{\exp(\beta_a)Z_a}{\sum_j\exp(\beta_j)Z_j},
&P(y)&=\nu_{y_1}\prod_{t=2}^D T_{y_{t-1},y_t}.
\end{aligned}
\label{app:theory-chain-construction}
\end{equation}
All probabilities are positive, and
\begin{equation}
\nu_aT_{ab}
=\frac{\exp(\beta_a+\beta_b+s_as_b)}{\sum_j\exp(\beta_j)Z_j}
=\nu_bT_{ba}
\end{equation}
proves stationarity and reversibility.
Choose $U_a=(s_a,\log Z_a,0,\ldots,0)$, $b_a=\beta_a$, and
\begin{equation}
\phi(h)=
\begin{cases}
(0,1,0,\ldots,0),&\text{empty prefix},\\
(s_{y_{\mathrm{last}}},0,0,\ldots,0),&\text{nonempty prefix}.
\end{cases}
\label{app:theory-easy-features}
\end{equation}
Here $y_{\mathrm{last}}$ is the state of the most recently revealed variable.
The empty-prefix output equals $\nu$, and subsequent outputs equal the required rows of $T$.
The Markov property and reversibility make these the full-prefix conditionals in both chain directions, so this shared predictor attains $H(P)/D$ with two features.
Other members of $\mathcal F_C$ are not restricted to reading one prefix variable.

\paragraph{Shared-output rank constraint.}
Stack features for $N$ histories into $\Phi\in\mathbb R^{N\times C}$ and predictions into $Q\in\mathbb R^{N\times V}$.
With $H_n=I_n-\mathbf 1_n\mathbf 1_n^{\mathsf T}/n$ and row log-normalizers $c$, the entrywise logarithm satisfies
\begin{equation}
\begin{aligned}
\log Q&=\Phi U^{\mathsf T}+\mathbf 1_Nb^{\mathsf T}-c\mathbf 1_V^{\mathsf T},\\
H_N\log QH_V&=(H_N\Phi)(U^{\mathsf T}H_V),
\qquad \operatorname{rank}(H_N\log QH_V)\le C.
\end{aligned}
\label{app:theory-shared-rank}
\end{equation}
Double centering removes shared biases and row normalizers, yielding a constraint on log-probability tables.

\paragraph{Lemma A.1 (centered logarithmic kernel).}
For distinct positive $x_1,\ldots,x_V$ and distinct positive $y_1,\ldots,y_V$, let $F_{ij}=\log(1+x_i y_j)$.
Then $\operatorname{rank}(H_VFH_V)=V-1$.

\paragraph{Proof.}
If the rank were smaller, there would be a nonzero $z\in\mathbf 1_V^\perp$ with $Fz=c\mathbf 1_V$.
Then
\begin{equation}
f(x)=\sum_j z_j\log(1+xy_j)-c,
\qquad f'(x)=\sum_j\frac{z_j}{x+1/y_j}
\end{equation}
has $V$ distinct positive zeros, so $f'$ has at least $V-1$ by Rolle's theorem.
Multiplying $f'$ by $\prod_j(x+1/y_j)$ gives a polynomial of degree at most $V-2$, since $\sum_jz_j=0$.
The denominator has no positive root, so this polynomial must vanish identically.
Evaluating it at each $-1/y_j$ forces $z_j=0$, since the poles are distinct, a contradiction.
The centering matrices already bound the rank above by $V-1$, proving equality.

\paragraph{A difficult two-step conditional with finite parameters.}
In Eq.~\eqref{app:theory-chain-construction}, take $s_1=0$, $s_2=1$, $\beta_1=\beta_2=0$, and $\beta_a=-M$ for $a\ge3$, keeping the remaining $s_a$ distinct and fixed.
For $K=T^2$, multiplication gives
\begin{equation}
\log K_{ac}
=\beta_c-\log Z_a+
\log\left[\sum_b\frac{\exp(\beta_b)}{Z_b}
\exp\bigl(s_b(s_a+s_c)\bigr)\right].
\label{app:theory-two-step-log}
\end{equation}
Double centering removes the first two terms.
As $M\to\infty$, $Z_1\to2$, $Z_2\to1+e$, and summands with $b\ge3$ vanish, so the remaining matrix converges entrywise to
\begin{equation}
-\log2+\log\left(1+\frac{2}{1+e}\exp(s_a+s_c)\right).
\end{equation}
Apply Lemma A.1 with $x_a=\exp(s_a)$ and $y_c=2\exp(s_c)/(1+e)$.
The centered limit has a nonzero minor of order $V-1$, which remains nonzero for sufficiently large finite $M$ by continuity.
Fix such a finite $M$; all probabilities remain positive and
\begin{equation}
\operatorname{rank}(H_V\log KH_V)=V-1.
\label{app:theory-difficult-rank}
\end{equation}
The limit only proves existence of finite parameters, without taking logarithms of zero-probability transitions.

\paragraph{Strictly positive conditional KL gap.}
An order beginning with $(\rho_1,\rho_3)$ requires $P(Y_3=c\mid Y_1=a)=K_{ac}$ at its second step.
Let $\mathcal Q_C$ contain the prediction tables $Q_{ac}$ realizable by $\mathcal F_C$ on these $V$ target-prefix histories, and define
\begin{equation}
\ell_C:=\inf_{Q\in\mathcal Q_C}
\sum_a\nu_a\operatorname{KL}(K_{a,:}\Vert Q_{a,:}).
\label{app:theory-local-kl-gap}
\end{equation}
We claim $\ell_C>0$.
Otherwise there is a sequence $Q^{(n)}\in\mathcal Q_C$ whose weighted KL tends to zero.
Since every $\nu_a>0$, each row KL tends to zero, and Pinsker's inequality gives $Q^{(n)}\to K$ entrywise.
Strict positivity of $K$ makes the entrywise logarithm continuous at $K$, hence $H_V\log Q^{(n)}H_V\to H_V\log KH_V$.
Every matrix in this sequence has rank at most $C$ by Eq.~\eqref{app:theory-shared-rank}.
The set of matrices of rank at most $C$ is closed, because all minors of order $C+1$ vanish on it.
The limit would therefore have rank at most $C\le V-2$, contradicting Eq.~\eqref{app:theory-difficult-rank}.
This proves a positive KL gap, without assuming that the infimum is attained.

\paragraph{From conditional KL to any-order risk.}
The chain rule for KL gives
\begin{equation}
\operatorname{KL}(P\Vert Q_{q,\pi})
=\sum_{t=1}^D\mathbb E_P
\operatorname{KL}\left(P(X_{\pi_t}\mid X_{\pi_{<t}})
\,\Vert\,q(\cdot\mid h_{\pi,t})\right).
\end{equation}
All terms are nonnegative, and a uniform permutation begins with $(\rho_1,\rho_3)$ with probability $1/[D(D-1)]$.
On that event, its second conditional KL is at least $\ell_C$ after averaging over $Y_1\sim\nu$.
Consequently, for every shared predictor,
\begin{equation}
\mathbb E_\Pi R(q,\Pi)\ge\frac{H(P)}D+\Gamma_C,
\qquad \Gamma_C=\frac{\ell_C}{D^2(D-1)}>0.
\label{app:theory-ao-gap}
\end{equation}
Taking the infimum preserves this bound; the explicit two-feature predictor establishes equality with $H(P)/D$ in both chain directions and completes Theorem A.1.
Equation~\eqref{eq:order-capacity-gap} therefore holds with $\Delta_C=\Gamma_C$.

This is a representational result for Eq.~\eqref{app:theory-model-class}, not a lower bound on arbitrary network parameter counts or a training-speed guarantee.
A subclass with the same output interface inherits the lower bound, while its chain-order upper bound additionally requires that it implement Eq.~\eqref{app:theory-easy-features}.
The theorem does not make every fixed random order worse or quantify the size of the gap in the experimental datasets.

\subsection{Graph alignment and conditional spectral recovery}
\label{app:spectral-recovery-proof}

We write $\widehat A$ for the empirical matrix $A$ defined by the exposure-corrected estimator in Appendix~\ref{app:method-exposure}, Eq.~\eqref{app:method-exposure-average}.
Its reciprocal graph $\widehat W$ is symmetric and nonnegative, with zero diagonal.
The following condition connects that graph to the same axis $\rho$ certified by Theorem A.1.

\paragraph{Assumption A.1 (alignment with the beneficial axis).}
Let $W_\rho$ be the unweighted path adjacency matrix on consecutive pairs $(\rho_t,\rho_{t+1})$.
Assume
\begin{equation}
\widehat W=\alpha W_\rho+\beta(\mathbf 1\mathbf 1^{\mathsf T}-I)+E,
\qquad \alpha>0,\quad\beta\ge0,
\label{app:theory-graph-alignment}
\end{equation}
where $E$ is symmetric with zero diagonal and $\widehat W$ remains nonnegative.
For any symmetric weight matrix $B$, write $L(B)=\operatorname{diag}(B\mathbf 1)-B$, and set $L_\rho=L(W_\rho)$ and $\eta=\|L(E)\|_{\mathrm{op}}$.
The substantive assumption is the bound on $\eta/\alpha$ below; $\alpha,\beta$ are graph coefficients unrelated to the categorical biases $\beta_a$.

\paragraph{Theorem A.2 (recovery of the chain axis).}
Define
\begin{equation}
(v_\rho)_{\rho_t}=-\sqrt{\frac2D}\cos\left(\frac{\pi(t-\tfrac12)}D\right),
\quad t=1,\ldots,D,
\label{app:theory-path-vector}
\end{equation}
and the positive spectral and coordinate gaps
\begin{equation}
\begin{aligned}
g_\rho&=\min\{\lambda_2(L_\rho),\lambda_3(L_\rho)-\lambda_2(L_\rho)\},\\
m_\rho&=\min_{t<D}\bigl((v_\rho)_{\rho_{t+1}}-(v_\rho)_{\rho_t}\bigr).
\end{aligned}
\end{equation}
Under Assumption A.1, if
\begin{equation}
\frac{\eta}{\alpha}<\min\left\{\frac{g_\rho}{4},
\frac{m_\rho g_\rho}{8\sqrt2}\right\},
\label{app:theory-spectral-threshold}
\end{equation}
then $\widehat L=L(\widehat W)$ has a simple Fiedler eigenvalue, and sorting a unit Fiedler vector $\widehat v$ recovers $\rho$ up to reversal.

\paragraph{Proof.}
The path eigenvalues are $\lambda_{k+1}(L_\rho)=2(1-\cos(k\pi/D))$ for $k=0,\ldots,D-1$.
Substitution verifies that Eq.~\eqref{app:theory-path-vector} is its unit, zero-mean Fiedler vector, with coordinates strictly increasing along $\rho$.
Thus both gaps are positive, and the path is recovered exactly, as in spectral seriation~\citep{atkins1998spectral,fogel2015convex}.
Linearity gives
\begin{equation}
\widehat L=L_0+L(E),
\qquad L_0=\alpha L_\rho+\beta(DI-\mathbf 1\mathbf 1^{\mathsf T}).
\end{equation}
On $\mathbf 1^\perp$, the background adds $\beta D$ times the identity, preserving eigenvectors and giving $\lambda_k(L_0)=\alpha\lambda_k(L_\rho)+\beta D$ for $k\ge2$.
Its Fiedler eigenvalue is separated from zero and the remaining spectrum by at least $\alpha g_\rho$.
Weyl's inequality and $\eta<\alpha g_\rho/4$ ensure that $\widehat L$ is connected and its Fiedler eigenvalue remains simple.

Projecting the perturbed eigenvector equation onto $v_\rho^\perp$, the inverse of $L_0-\lambda_2(\widehat L)I$ on that subspace has norm at most $(\alpha g_\rho-\eta)^{-1}$.
Hence the acute angle $\theta$ between the eigenvector lines satisfies $\sin\theta\le\eta/(\alpha g_\rho-\eta)\le2\eta/(\alpha g_\rho)$, the standard Davis--Kahan argument~\citep{yu2015useful}.
After choosing a common sign $s\in\{-1,1\}$,
\begin{equation}
\|s\widehat v-v_\rho\|_\infty
\le\|s\widehat v-v_\rho\|_2
\le\sqrt2\sin\theta
\le\frac{2\sqrt2\eta}{\alpha g_\rho}<\frac{m_\rho}{4}.
\label{app:theory-coordinate-error}
\end{equation}
Every consecutive coordinate gap remains greater than $m_\rho/2$, proving recovery up to reversal.

Uniform background does not change the exact ordering; nonuniform background must be included in $E$.
If the eigensolver contributes an additional sign-aligned coordinate error $\varepsilon_{\mathrm{eig}}$, the same proof applies when
\begin{equation}
\eta<\frac{\alpha g_\rho}{4},
\qquad
\frac{2\sqrt2\eta}{\alpha g_\rho}+\varepsilon_{\mathrm{eig}}<\frac{m_\rho}{2}.
\label{app:theory-numerical-error}
\end{equation}
These are deterministic conditions, not finite-sample guarantees that a trained Transformer satisfies them.
In particular, low predictive loss alone need not identify raw attention, because value representations or other computational paths may compensate for its weights.
More attention samples may reduce estimation error but do not automatically remove structural bias or checkpoint drift.

\subsection{Attention-based orientation}
\label{app:attention-orientation}

Let $\pi^{\mathrm{raw}}=\operatorname{argsort}_{\mathrm{asc}}(\widehat v)$.
For target-to-observed $A$, define $S_A(\pi)=\sum_{s<t}A_{\pi_t,\pi_s}$ and let $B_{ts}=A_{\pi_t^{\mathrm{raw}},\pi_s^{\mathrm{raw}}}$.

\paragraph{Proposition A.1 (triangular and antisymmetric forms).}
The forward and reverse scores are
\begin{equation}
S_A(\pi^{\mathrm{raw}})=\sum_{t>s}B_{ts},
\qquad
S_A(\operatorname{rev}(\pi^{\mathrm{raw}}))=\sum_{t<s}B_{ts}.
\label{app:theory-triangular-scores}
\end{equation}
Writing $M_A=\sum_{i\ne j}A_{ij}$ and $A^-=(A-A^{\mathsf T})/2$, every order satisfies
\begin{equation}
\begin{aligned}
S_A(\pi)+S_A(\operatorname{rev}(\pi))&=M_A,\\
S_A(\pi)-S_A(\operatorname{rev}(\pi))
&=\sum_{s<t}(A_{\pi_t,\pi_s}-A_{\pi_s,\pi_t})
=2\sum_{s<t}A^-_{\pi_t,\pi_s}.
\end{aligned}
\label{app:theory-antisymmetric-score}
\end{equation}

\paragraph{Proof.}
Reversal swaps the earlier and later member of each unordered pair, interchanging its lower- and upper-triangular entries.
Adding the scores counts both directed entries once, while subtracting them gives their signed difference.
Summing over pairs proves both identities.

Writing $\pi_T=\pi^{\mathrm{spec}}$ for the Spectral Teacher order defined in the main text, the Teacher selects
\begin{equation}
\pi_T\in\operatorname*{arg\,max}_{\pi\in\{\pi^{\mathrm{raw}},\operatorname{rev}(\pi^{\mathrm{raw}})\}}
S_{\widehat A}(\pi).
\label{app:theory-teacher-selection}
\end{equation}
Diagonal entries are excluded, and symmetric $A$ gives a tie.
The score compares prefix-compatible attention on the existing matrix, without new backbone evaluations or an identity relating the score to NLL.
The reciprocal graph is invariant under $A\mapsto A^{\mathsf T}$, whereas the score difference changes sign.
For Theorem A.1, both recovered directions are already beneficial, so any maximizer, including either tied candidate, preserves the capacity advantage.
For distributions with only one beneficial direction, an additional relation between that direction and the attention score would be needed.

\subsection{Priority normalization and preservation by the Order MLP}
\label{app:theory-distillation}

Using the main text's convention that earlier variables have larger priorities, define
\begin{equation}
z^{\mathrm{early}}=
\begin{cases}
-\widehat v,&\pi_T=\pi^{\mathrm{raw}},\\
\widehat v,&\pi_T=\operatorname{rev}(\pi^{\mathrm{raw}}),
\end{cases}
\qquad
r_i=\frac{z_i^{\mathrm{early}}-\min_jz_j^{\mathrm{early}}}
{\max_jz_j^{\mathrm{early}}-\min_jz_j^{\mathrm{early}}}.
\label{app:theory-priority-normalization}
\end{equation}
Theorem A.2 ensures distinct coordinates, so descending sorting of $r$ gives $\pi_T$.
With a unique score maximizer, the arbitrary Fiedler sign only swaps candidate names.
In a tie, selecting the lexicographically smaller current-ID sequence is one sign-invariant convention; either candidate remains certified.

\paragraph{Proposition A.2 (ranking fidelity).}
For the final Teacher target, let
\begin{equation}
\Delta_r=\min_{t<D}(r_{\pi_{T,t}}-r_{\pi_{T,t+1}})>0.
\end{equation}
Descending sorting of the final MLP output $\widehat r$ recovers $\pi_T$ if
\begin{equation}
\|\widehat r-r\|_\infty<\frac{\Delta_r}{2}.
\label{app:theory-mlp-margin}
\end{equation}
A sufficient condition on the same output vector is
\begin{equation}
\operatorname{MSE}(\widehat r,r)
:=D^{-1}\sum_i(\widehat r_i-r_i)^2
<\frac{\Delta_r^2}{4D}.
\label{app:theory-mlp-mse}
\end{equation}

\paragraph{Proof.}
For $e=\|\widehat r-r\|_\infty$, each consecutive difference is at least $\Delta_r-2e>0$, preserving the complete order.
The MSE condition follows from $\|\widehat r-r\|_\infty\le\|\widehat r-r\|_2=\sqrt{D\operatorname{MSE}(\widehat r,r)}$.

The margin uses the normalized $[0,1]$ scale, and the error must hold on the final graph used for the executable order.
Average MSE across windows alone does not imply this condition or establish convergence of distillation.

\subsection{Conditional risk guarantee}
\label{app:theory-risk-combination}

\paragraph{Theorem A.3 (preservation of the capacity advantage).}
Consider the same distribution and output class as Theorem A.1.
Suppose the actual attention graph satisfies Assumption A.1 and Eq.~\eqref{app:theory-spectral-threshold}, with an exact Fiedler vector or numerical error controlled by Eq.~\eqref{app:theory-numerical-error}.
Let $\pi_T$ follow Eq.~\eqref{app:theory-teacher-selection}, and either execute $\widehat\pi=\pi_T$ directly or obtain $\widehat\pi$ by descending sorting of an MLP output satisfying Eq.~\eqref{app:theory-mlp-margin}.
If the final predictor $\widehat q\in\mathcal F_C$ satisfies
\begin{equation}
R(\widehat q,\widehat\pi)
\le R_C^\star(P,\widehat\pi)+\varepsilon_{\mathrm{fit}},
\label{app:theory-final-fit}
\end{equation}
then
\begin{equation}
\begin{aligned}
\widehat\pi&\in\{\rho,\operatorname{rev}(\rho)\},\\
R(\widehat q,\widehat\pi)&\le\frac{H(P)}D+\varepsilon_{\mathrm{fit}},\\
R_{\mathrm{AO},C}^\star(P)-R(\widehat q,\widehat\pi)
&\ge\Gamma_C-\varepsilon_{\mathrm{fit}}.
\end{aligned}
\label{app:theory-final-risk-bound}
\end{equation}

\paragraph{Proof.}
Theorem A.2 recovers exactly the two chain directions, orientation selects one, and Proposition A.2 preserves it when the MLP is used.
Theorem A.1 gives $R_C^\star(P,\widehat\pi)=H(P)/D$; combining the fitting bound with its AO lower bound proves the result.
Both directions are certified, so no positive orientation margin is required.

Strict improvement requires $\varepsilon_{\mathrm{fit}}<\Gamma_C$ and compares the selected fixed-order risk with the uniform-order average in the same class.
The theorem first certifies beneficial orders independently of attention, then assumes graph alignment and controls recovery, distillation, and fitting.
It does not establish that native Transformer training meets these conditions, certify them for the reported datasets, or identify a uniquely optimal direction.

\section{Implementation Details}
\label{app:method-details}

\subsection{Attention Alignment and Exposure Correction}
\label{app:method-coordinates}

Attention matrices from different reveal orders must refer to the same ordered variable pairs before they are averaged.
Let $\sigma$ map original variable IDs to current IDs, so that $x_i^{\mathrm{cur}}=x_{\sigma^{-1}(i)}^{\mathrm{orig}}$.
This relabeling is fixed across examples; variable-aligned conditions follow it, whereas global conditions are unchanged.
For example $b$, let $\pi_b$ be its reveal permutation and $r_b(i)=\pi_b^{-1}(i)$ its one-based rank map.
Original IDs and semantic names are reserved for post-hoc interpretation.

The decoder input is $(\mathrm{start},x_{\pi_{b,1}},\ldots,x_{\pi_{b,D}})$: query position $t-1$ predicts $x_{\pi_{b,t}}$, while key position $s\geq1$ contains $x_{\pi_{b,s}}$.
Consequently, the target-to-observed view of raw $(D+1)\times(D+1)$ attention $R^{(b,\ell,h)}$ is $\widetilde a^{(b,\ell,h)}_{ts}=R^{(b,\ell,h)}_{t-1,s}$ for $1\leq t,s\leq D$.
The zero-based slice \texttt{R[:-1, 1:]} removes the start-state key and the unused final query, giving the visibility condition $s<t$.
An unshifted row index would associate the query with the previous observed variable rather than its prediction target.
We map both axes to current IDs before averaging:
\begin{equation}
a^{(b,\ell,h)}_{ij}=\widetilde a^{(b,\ell,h)}_{r_b(i),r_b(j)},
\qquad M^{(b)}_{ij}=\mathbb{I}\{r_b(j)<r_b(i)\}.
\label{app:method-alignment}
\end{equation}
The mask includes a pair only when the column variable was legally visible to the row target.
Removing the start key does not redistribute its original softmax mass over the retained keys.

\paragraph{Exposure correction.}
\label{app:method-exposure}
For $N$ examples, average all $N_L$ decoder layers and $N_H$ heads equally, then accumulate masked sums and legal-exposure counts:
\begin{align}
\bar a^{(b)}_{ij}&=\frac{1}{N_LN_H}\sum_{\ell,h}a^{(b,\ell,h)}_{ij},\notag\\
U_{ij}&=\sum_bM^{(b)}_{ij}\bar a^{(b)}_{ij},
\qquad C_{ij}=\sum_bM^{(b)}_{ij},\notag\\
A_{ij}&=\frac{U_{ij}}{\max(1,C_{ij})}\quad(i\ne j),
\qquad A_{ii}=0.
\label{app:method-exposure-average}
\end{align}
Zero-exposure entries are zero, and attention is detached before accumulation.
Dividing by $C_{ij}$ prevents frequently exposed pairs from receiving larger affinities merely because they occurred more often.
The correction removes unequal exposure frequency, not dependence on the backbone or reveal-context distribution; annealing windows therefore reflect the current order mixture.
These affinities remain model-induced relational signals rather than estimates of causal influence.
The two $D\times D$ accumulators span the task-specific windows in Appendix~\ref{app:protocol-schedules}.
Completed windows yield $W_{ij}=\sqrt{A_{ij}A_{ji}}$ with zero diagonal and $L=\operatorname{diag}(W\mathbf1)-W$; spectral recovery and orientation follow Appendices~\ref{app:spectral-recovery-proof} and~\ref{app:attention-orientation}.

\subsection{Order-MLP Parameterization}
\label{app:method-mlp}

The Spectral Teacher orients the Fiedler coordinates using Eq.~\eqref{eq:attention-orientation} and normalizes them to $[0,1]$ using Eq.~\eqref{app:theory-priority-normalization}.
The resulting target $\mathbf q^*=\mathbf r$ assigns larger priorities to variables generated earlier.
The student receives directed $A$, with population z-normalization over all $D^2$ entries, including the zero diagonal:
\begin{equation}
\begin{aligned}
\mu_A&=D^{-2}\sum_{i,j}A_{ij},\qquad A^{\mathrm{std}}=(A-\mu_A)/s_A,\\
s_A&=\max\left(10^{-6},\sqrt{D^{-2}\sum_{i,j}(A_{ij}-\mu_A)^2}\right).
\end{aligned}
\label{app:method-mlp-normalization}
\end{equation}

The flattened $A^{\mathrm{std}}$ passes through two fully connected GELU hidden layers of width 32 and a $D$-unit sigmoid output layer to produce priorities $\widehat{\mathbf q}_{\phi}$, giving the architecture $D^2\rightarrow32\rightarrow32\rightarrow D$ for every task.
Training and order extraction use
\begin{equation}
\mathcal L_{\mathrm{order}}=D^{-1}\sum_i(\widehat q_{\phi,i}-q_i^*)^2,
\qquad
\widehat\pi_{\phi}=\operatorname{argsort}_{\mathrm{descending}}(\widehat{\mathbf q}_{\phi}).
\label{app:method-mlp-objective}
\end{equation}
The $[0,1]$ outputs are priorities, not calibrated probabilities, and continuous targets retain relative spectral spacing.
Both $A$ and $\mathbf q^*$ are detached, so MSE updates only the MLP; each window produces one task-level order shared across examples.

\subsection{Training Lifecycle}
\label{app:method-lifecycle}

Table~\ref{app:protocol-schedule-table} reports three milestones: starting distillation, stopping distillation, and completing annealing.
For the current learned order $\widehat\pi_s$ available at update $s$,
\begin{equation}
q_s(\pi)=(1-p_s)\operatorname{Unif}(\mathfrak S_D)+p_s\delta_{\widehat\pi_s}(\pi),
\label{app:method-mixture}
\end{equation}
where $p_s$ gradually increases from zero to one during training.
During distillation, the backbone trains with random reveal orders while the Spectral Teacher supervises the Order MLP using aggregated attention.
After distillation, MLP parameters remain fixed, and learned-order use is gradually increased.
Within the configured refresh window, the MLP proposes candidate orders from new attention estimates for subsequent backbone updates.
After annealing completes, the final accepted permutation is reused for training and generation.

Generation with the final learned order requires neither MLP inference nor an eigensolve.

\section{Experimental Protocols}
\label{app:protocol-details}

\subsection{Datasets and Representations}
\label{app:protocol-data}

\paragraph{WikiText103 Seq80.}
The mechanism experiment uses raw WikiText103~\citep{merity2017pointer}, GPT-2 BPE, and an end-of-text token after each source record.
Training and official validation streams use unsigned 16-bit IDs and 80-token windows; the test split is excluded.
Each position is one variable, relabeled once with permutation seed 42.
This experiment studies order recovery and distillation under a short context; it does not train the model used for the OpenWebText zero-shot benchmark.

\paragraph{OpenWebText.}
OpenWebText~\citep{gokaslan2019openwebtext} uses the same tokenizer and document-end convention, separate unsigned 16-bit streams, and a document-level validation fraction $0.0005$ with split seed 2357.
Training uses 1,024-token windows, a 50,257-symbol vocabulary padded to 50,304 outputs, and relabeling seed 42.
The 400k budget counts repeated token presentations, not distinct corpus tokens.

\paragraph{H3WB masked-I2D.}
Each of 133 three-dimensional joint targets is conditioned on two-dimensional keypoints and visibility indicators~\citep{zhu2023h3wb}.
Targets are pelvis-relative; observed coordinates retain official image coordinates before normalization.
Official training sequences are partitioned by group ID with seed 1337, targeting fractions $0.1$ for validation and $0.1$ for discovery; preserving groups permits small deviations from those fractions.
Training-partition coordinate means and standard deviations are fixed for validation and the released 10,000-pose test set.
The ordinary-training window protocol collects attention from training batches; the existence of a separate prepared discovery partition does not mean this protocol draws its attention windows from that partition.
Corruption mixes independent masking (weight $0.4$, per-keypoint probability $0.25$) with whole-face, left-hand, and right-hand masks (weight $0.2$ each).

\paragraph{CENSUS.}
The processed 1990 U.S. Census table~\citep{meek2001uscensus} has 2,458,285 rows and 68 categorical variables after removing the case identifier.
Each column receives a fixed sorted-unique vocabulary of dense zero-based IDs.
The already randomized table is split contiguously into 2,212,456 training, 122,914 validation, and 122,915 test rows.
Semantic names do not enter order learning.
Representative permuted H3WB/CENSUS runs use relabeling seed 0; canonical baselines retain source indexing.
The source-domain category vocabulary is fixed during preparation, and mapping predictions back to original IDs is an evaluation operation.

\subsection{Backbones and Optimization}
\label{app:protocol-optimization}

Table~\ref{app:protocol-backbone-table} specifies backbone sizes and schedules.
All use zero dropout, AdamW weight decay $0.1$ on eligible parameters, gradient clipping $1.0$, and cosine learning-rate decay after warm-up.
Adam coefficients are $(0.9,0.99)$ for Seq80 and $(0.9,0.95)$ otherwise.
Language decoders use learned variable-position embeddings and target-position conditioning.
H3WB adds one bidirectional condition-encoder layer and diagonal-Gaussian joint outputs with log scale clamped to $[-7,5]$; CENSUS uses column-specific categorical outputs.

\begin{table}[t]
\centering
\small
\caption{Backbone settings: architecture denotes decoder layers / heads / hidden width, and batch counts examples per optimizer update.}
\label{app:protocol-backbone-table}
\begin{tabular}{@{}lrrrr@{}}
\toprule
Setting & Seq80 & OpenWebText & H3WB & CENSUS \\
\midrule
Ordered variables $D$ & 80 & 1,024 & 133 & 68 \\
Decoder architecture & $4/8/384$ & $12/10/640$ & $4/8/256$ & $2/4/128$ \\
Global batch & 512 & 480 & 160 & 512 \\
Backbone updates & 50,000 & 400,000 & 100,000 & 50,000 \\
Peak learning rate & $10^{-3}$ & $6\times10^{-4}$ & $3\times10^{-4}$ & $3\times10^{-4}$ \\
Minimum learning rate & $10^{-4}$ & $6\times10^{-5}$ & $3\times10^{-5}$ & $3\times10^{-5}$ \\
Warm-up updates & 0 & 2,000 & 5,000 & 1,000 \\
\bottomrule
\end{tabular}
\end{table}

The OpenWebText backbone has 130,336,512 parameters.
The OpenWebText OrderGPT and random-baseline runs use six RTX 5880 Ada Generation GPUs.
Both use a global batch of 480, totaling 196.608 billion token presentations over 400k updates.
All protocols use BF16, with compiled execution for language and CENSUS.

All tasks use the Order-MLP architecture described in Appendix~\ref{app:method-mlp}.
Order MLP uses AdamW with learning rate $2\times10^{-4}$, weight decay $0.01$, gradient clipping $1.0$, and zero dropout.
Adam coefficients are $(0.9,0.99)$ for language and $(0.9,0.999)$ for H3WB/CENSUS.
Its input / hidden / output widths are $6{,}400\rightarrow32\rightarrow32\rightarrow80$ (Seq80), $1{,}048{,}576\rightarrow32\rightarrow32\rightarrow1{,}024$ (400k OpenWebText), $17{,}689\rightarrow32\rightarrow32\rightarrow133$ (H3WB), and $4{,}624\rightarrow32\rightarrow32\rightarrow68$ (CENSUS).

\subsection{Attention Windows and Order-Learning Milestones}
\label{app:protocol-schedules}

Table~\ref{app:protocol-schedule-table} lists the attention windows and the three order-learning milestones in backbone optimizer updates.
Window sizes count examples across optimizer updates, rather than micro-batches.
All Spectral Teacher and OrderGPT experiments use the attention-based orientation rule in Eq.~\eqref{eq:attention-orientation}.

\begin{table}[t]
\centering
\small
\caption{Attention windows and three order-learning milestones.}
\label{app:protocol-schedule-table}
\begin{tabular}{@{}lrrrr@{}}
\toprule
Setting & Seq80 & OpenWebText & H3WB & CENSUS \\
\midrule
Examples per attention window & 1,024 & 2,400 & 1,280 & 1,024 \\
Updates per window & 2 & 5 & 8 & 2 \\
Start distillation & 10k & 4k & 15k & 5k \\
Stop distillation & 18k & 40k & 40k & 10k \\
Annealing complete & 32k & 60k & 60k & 20k \\
\bottomrule
\end{tabular}
\end{table}

\subsection{Baselines and Comparison Scope}
\label{app:protocol-baselines}

Canonical AR uses L2R in language and source indexing in H3WB/CENSUS; Fixed-random AR (throughout) uses one presampled permutation for training and inference; AO-GPT samples a fresh permutation per training example.
Random and canonical language inference use the same AO-GPT checkpoint.
Spectral Teacher uses the attention-oriented spectral candidate, whereas OrderGPT sorts the MLP priorities.
The two 400k OpenWebText fixed-random controls share their terminal permutation and coordinate mapping.
Fixed-random AR + curriculum uses the same order-use probability $p_s$ as OrderGPT in Eq.~\eqref{app:method-mixture}, with one presampled random permutation replacing the learned order.
This control tests whether the same schedule with an arbitrary terminal order reproduces the gains of OrderGPT.
Within each domain, internal comparisons match the backbone capacity and data processing as well as the backbone-update budget.
Matched backbone updates and token presentations do not imply identical hardware, elapsed time, or total compute including order discovery and MLP training.
Internal OpenWebText runs omit language-model EMA and prescribed L2R training mixtures; published AO-GPT rows retain those protocol differences~\citep{xue2025any} and serve as external context.

\subsection{Evaluation and Statistical Units}
\label{app:protocol-evaluation}

\paragraph{Language.}
Perplexity is $\exp\{(\sum_n\ell_n)/N_{\mathrm{tokens}}\}$ for natural-log token losses.
The fixed-record evaluator maps 1,024-token benchmark records into checkpoint current IDs, scores all targets including the first, and accumulates losses in float64.
Inference uses random, original-L2R, or frozen learned order as specified, without benchmark-side order discovery.
For multiple random orders, losses are averaged before exponentiation, not perplexities afterward.
LAMBADA reports token perplexity, not last-word accuracy; Seq80 validation is a separate short-context experiment.

\paragraph{H3WB.}
Teacher-forced NLL is averaged per three-dimensional joint; closed-loop evaluation generates all 133 joints by successive Gaussian means and reports MPJPE in millimeters.
Whole/Body use pelvis alignment, Face-local uses nose alignment, and Hand-local uses the corresponding wrist.
The full 10,000-pose test protocol uses BF16, batch 256, and evaluation-order seed 20260713.
Checkpoint selection uses the configured validation criterion after the method's selection boundary, which is 60k for OrderGPT.
Teacher-forced NLL and free-running MPJPE measure different outcomes.

\paragraph{CENSUS.}
Test NLL is averaged per variable on the held-out slice.
Cardinality evaluation uses 1,000 shared queries with 5--11 predicates (seed 1234), exact counts on all 2,458,285 rows, and 10,000 particles per query.
Predicate IDs are remapped for permuted checkpoints without changing query meaning.
Fixed-order models use their saved permutation; the random-order model draws one per particle and keeps it for all 68 steps.
Estimated cardinality is the ceiling of mean particle weight times table size.
For positive counts, Q-error is $\max(\widehat c/c,c/\widehat c)$; P50, P99, and maximum are the workload median, linearly interpolated $0.99$ quantile, and maximum.
Published references retain their original protocols~\citep{yang2019deepcardinality,liang2020variableskipping}.

\paragraph{Replicates.}
Main-table H3WB/CENSUS entries report means and sample standard deviations across independently trained models, with replicate counts in the captions.
Q-error percentiles and maxima are computed within each run before aggregation.
Relabeling, reveal-order, and particle-sampling randomness do not constitute additional training replicates.
Appendix curves and order visualizations show individual runs and provide no across-run uncertainty estimate or significance test.

\paragraph{Fixed-checkpoint efficiency.}
\label{app:protocol-efficiency}
Seq80 efficiency is measured by replaying an 18k-update checkpoint on one RTX 4090 with BF16 and compiled execution.
Both methods aggregate 1,024 examples and refresh the order every two backbone updates.
Timing uses synchronized measurements over 200 updates after a 100-update warm-up, with paired random baselines.
These measurements concern per-update execution rather than complete-training wall time.

\section{Additional Results and Analyses}
\label{app:result-overview}

\subsection{Training Dynamics}
\label{app:result-training}

\begin{figure}[t]
\centering
\includegraphics[width=\textwidth]{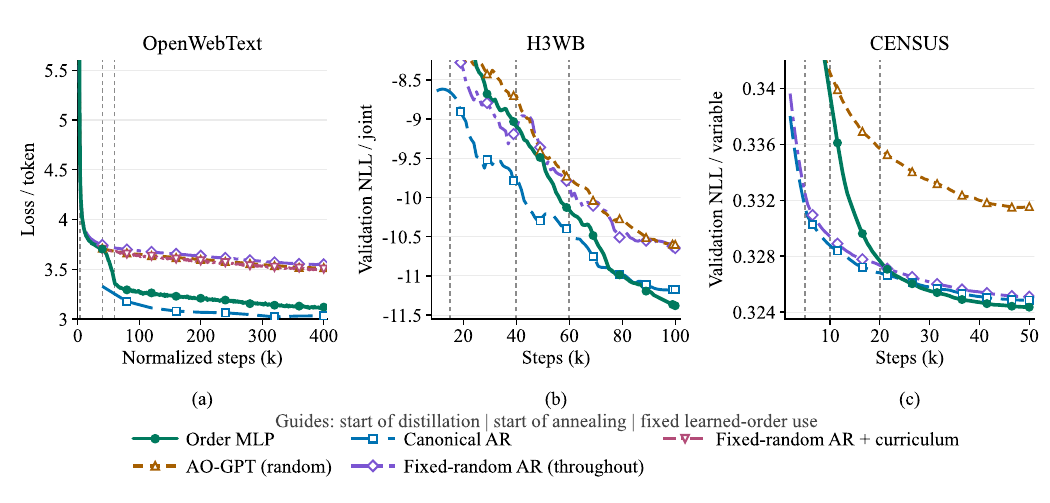}
\caption{Loss trajectories: \textbf{(a)} OpenWebText training token NLL ; \textbf{(b)} H3WB validation NLL per joint; \textbf{(c)} CENSUS validation NLL per variable.
Vertical guides mark curriculum transitions.
The OpenWebText Fixed-random AR + curriculum curve follows the same schedule as OrderGPT.
Internal curves represent individual runs and are averaged or smoothed for display.}
\label{app:result-loss-figure}
\end{figure}

Figure~\ref{app:result-loss-figure} shows training dynamics as learned-order use increases.
Order MLP reaches a lower final training loss than the random-order and fixed-random baselines on OpenWebText.
Its validation trajectory separates from the random-order baseline on H3WB and CENSUS.

\subsection{Structure of the Learned H3WB Order}
\label{app:result-h3wb-structure}

\begin{figure}[t]
\centering
\includegraphics[width=\textwidth]{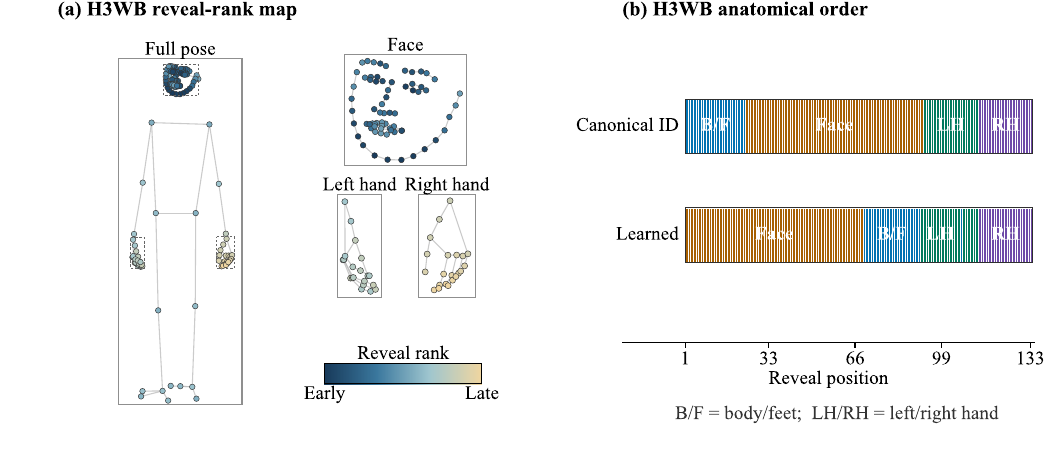}
\caption{Post-hoc interpretation of one learned H3WB order.
\textbf{(a)} Reveal rank on a representative training-set 2D pose, with face and hand enlargements; darker points appear earlier.
\textbf{(b)} Anatomical regions along canonical and learned orders.
The pose and connecting lines provide display context, not an input-specific prediction, learned dependency edges, or the evaluation bone set.
The pose, canonical identities, and anatomical labels do not enter order learning, direction selection, or model selection.}
\label{app:result-h3wb-figure}
\end{figure}

The order in Figure~\ref{app:result-h3wb-figure} places all 68 facial joints first, followed by 22 of the 23 body-and-foot joints and then the hands, with the remaining body joint interleaved in the left-hand span.
This coarse face-to-body-to-hands traversal differs from the dataset's body/feet-to-face-to-hands indexing, while the interleaving shows that the learned order is not a strict concatenation of anatomical groups.

\subsection{CENSUS Locality and Representative Neighborhoods}
\label{app:result-census-structure}

For rank $r_\pi(i)$ of variable $i$ and a fixed nonnegative symmetric affinity matrix $W$ with positive off-diagonal weight, define
\begin{equation}
  d_W(\pi)
  =
  \frac{\sum_{i<j}W_{ij}|r_\pi(i)-r_\pi(j)|}
       {\sum_{i<j}W_{ij}}.
  \label{app:result-locality-definition}
\end{equation}
Under a uniform random permutation independent of $W$, the expected distance is $(D+1)/3$, giving $23$ for the 68 CENSUS variables.

The representative audit constructs $W$ from a frozen random-order AO-GPT checkpoint on 2,048 training examples, averaging its two layers and four heads with exposure correction.
After aligning variable identities, the same matrix is evaluated under canonical indexing and the saved, frozen Order-MLP permutation, without recomputing attention under that permutation.
Weighted separations are $17.91$ for the learned order and $23.17$ for canonical indexing, compared with the random expectation of $23.00$.
The graph and saved permutation come from different checkpoints but use the same task and training split, so this comparison is a structural diagnostic rather than an independent test of generalization.

\begin{table}[t]
\centering
\small
\caption{Representative CENSUS neighborhoods in the audited order; positions are one-based.
Column names and semantic groups are post-hoc annotations unavailable to the order learner, and the windows need not form a disjoint partition.}
\label{app:result-census-neighborhoods}
\setlength{\tabcolsep}{4pt}
\renewcommand{\arraystretch}{1.15}
\begin{tabular}{@{}>{\raggedright\arraybackslash}p{0.24\textwidth}>{\raggedright\arraybackslash}p{0.72\textwidth}@{}}
\toprule
Neighborhood or check & Observed fields and positions \\
\midrule
Service/military, 1--10 & \texttt{iOthrserv}, \texttt{iMay75880}, \texttt{iKorean}, \texttt{iWWII}, \texttt{iRvetserv}, \texttt{dYrsserv}, \texttt{iSept80}, \texttt{iFeb55}, \texttt{iVietnam}, \texttt{iMilitary}. \\
Family, 13--14, 21--22, 28--29 & \texttt{iRelat1}--\texttt{iFertil}; \texttt{iRownchld}--\texttt{iMarital}; \texttt{iRspouse}--\texttt{iRrelchld}. \\
Occupation/labor, 31--39 & \texttt{dOccup}, \texttt{dHour89}, \texttt{dIndustry}, \texttt{iMobility}, \texttt{iTmpabsnt}, \texttt{dDepart}, \texttt{dHours}, \texttt{iYearwrk}, \texttt{iRemplpar}. \\
Language/origin, 57--62 & \texttt{dAncstry2}, \texttt{iCitizen}, \texttt{dAncstry1}, \texttt{iSubfam1}, \texttt{iEnglish}, \texttt{dPOB}. \\
Pairwise distances & \texttt{iRvetserv}--\texttt{dYrsserv}: $1$; \texttt{dOccup}--\texttt{dIndustry}: $2$; \texttt{dIncome1}--\texttt{dRearning}: $24$. \\
\bottomrule
\end{tabular}
\end{table}

Table~\ref{app:result-census-neighborhoods} shows a service-and-military prefix and shorter family neighborhoods, alongside interleaved mobility and family fields in the occupation and language windows.
The income pair remains 24 positions apart: improved global locality does not force every related pair to be adjacent.

\FloatBarrier
\section{Extended Related Work}
\label{app:related-overview}

\paragraph{Masked diffusion and any-order autoregression.}
Deep NADE shares parameters across autoregressive decompositions with different variable orders~\citep{uria2016neural}.
AO-GPT connects masked-diffusion and any-order objectives while separating their formulation from a decoder-only implementation~\citep{xue2025any}.
The connection matches distributions over context sizes, observed subsets, and prediction targets; it does not equate conditional parameterizations or sampling trajectories.
A uniform permutation and uniformly sampled prefix length induce a uniform observed subset at each size and a uniform next target among unrevealed variables.
Matching these weights explains the objective correspondence even when the decoder retains the prefix's presentation order.
OrderGPT further selects one reusable task-level factorization instead of continuing to average training over all orders.

\paragraph{Adaptive paths and learned policies.}
\citet{pmlr-v267-kim25ah} establish difficult infilling problems in constructed masked-diffusion settings and show that adaptive decoding can avoid some of them.
Their conditional-inference complexity result is distinct from our shared-output-feature capacity separation.
Path Planning selects state-dependent denoising updates, including revisions to previously unmasked variables~\citep{peng2025path}.
LO-ARM jointly learns a state-dependent probabilistic autoregressive order policy through a variational likelihood objective~\citep{pmlr-v267-wang25cn}.
OrderGPT instead distills shared dependency structure into a fixed permutation, eliminating generation-time policy decisions while sacrificing per-example adaptation.
Our experiments do not compare its performance against adaptive planners.

\paragraph{Attention and spectral structure.}
Affinity from Attention learns semantic affinities from Transformer attention using segmentation pseudo labels~\citep{ru2022affinity}.
OrderGPT instead aligns reveal contexts, corrects pair exposure, and combines reciprocal weights without semantic-affinity or canonical-order supervision.
The resulting graph remains conditioned on the trained backbone and the reveal distribution used for aggregation.
Spectral seriation reconstructs linear arrangements from structured similarities~\citep{atkins1998spectral}; its connection to 2-SUM motivates the locality objective~\citep{fogel2015convex}.
These results do not establish that arbitrary Transformer attention meets the required structure or that spectral sorting minimizes NLL.
We therefore treat attention-graph alignment as an explicit condition, separate from the capacity theorem, and interpret directed-attention orientation as an endpoint preference rather than causal identification.

\end{document}